\documentclass[10pt,twocolumn,letterpaper]{article}

\usepackage[pagenumbers]{cvpr} 

\usepackage{graphicx}
\usepackage{amsmath}
\usepackage{amssymb}
\usepackage{booktabs}
\usepackage{gensymb}
\usepackage{multirow}

\usepackage[pagebackref,breaklinks,colorlinks]{hyperref}

\usepackage[accsupp]{axessibility} 

\usepackage[capitalize]{cleveref}
\crefname{section}{Sec.}{Secs.}
\Crefname{section}{Section}{Sections}
\Crefname{table}{Table}{Tables}
\crefname{table}{Tab.}{Tabs.}

\def\cvprPaperID{4643} 
\def\confName{CVPR}
\def\confYear{2023}

\begin{document}

\title{OcTr: Octree-based Transformer for 3D Object Detection}

\author{Chao Zhou\textsuperscript{\rm 1,2}, Yanan Zhang\textsuperscript{\rm 1,2}, Jiaxin Chen\textsuperscript{\rm 2}, Di Huang\textsuperscript{\rm 1,2,3}\thanks{indicates the corresponding author.}\\
\textsuperscript{\rm 1}State Key Laboratory of Software Development Environment, Beihang University, Beijing, China\\
\textsuperscript{\rm 2}School of Computer Science and Engineering, Beihang University, Beijing, China\\
\textsuperscript{\rm 3}Hangzhou Innovation Institute, Beihang University, Hangzhou, China\\
{\tt\small \{zhouchaobeing, zhangyanan, jiaxinchen, dhuang\}@buaa.edu.cn}
}

\maketitle

\begin{abstract}
   A key challenge for LiDAR-based 3D object detection is to capture sufficient features from large scale 3D scenes especially for distant or/and occluded objects. Albeit recent efforts made by Transformers with the long sequence modeling capability, they fail to properly balance the accuracy and efficiency, suffering from inadequate receptive fields or coarse-grained holistic correlations. In this paper, we propose an \textbf{Oc}tree-based \textbf{Tr}ansformer, named \textbf{OcTr}, to address this issue. It first constructs a dynamic octree on the hierarchical feature pyramid through conducting self-attention on the top level and then recursively propagates to the level below restricted by the octants, which captures rich global context in a coarse-to-fine manner while maintaining the computational complexity under control. Furthermore, for enhanced foreground perception, we propose a hybrid positional embedding, composed of the semantic-aware positional embedding and attention mask, to fully exploit semantic and geometry clues. Extensive experiments are conducted on the Waymo Open Dataset and KITTI Dataset, and OcTr reaches newly state-of-the-art results.

\end{abstract}

\section{Introduction}
\label{sec:intro}

\begin{figure}[t]
    \centering
    \includegraphics[width=0.47\textwidth]{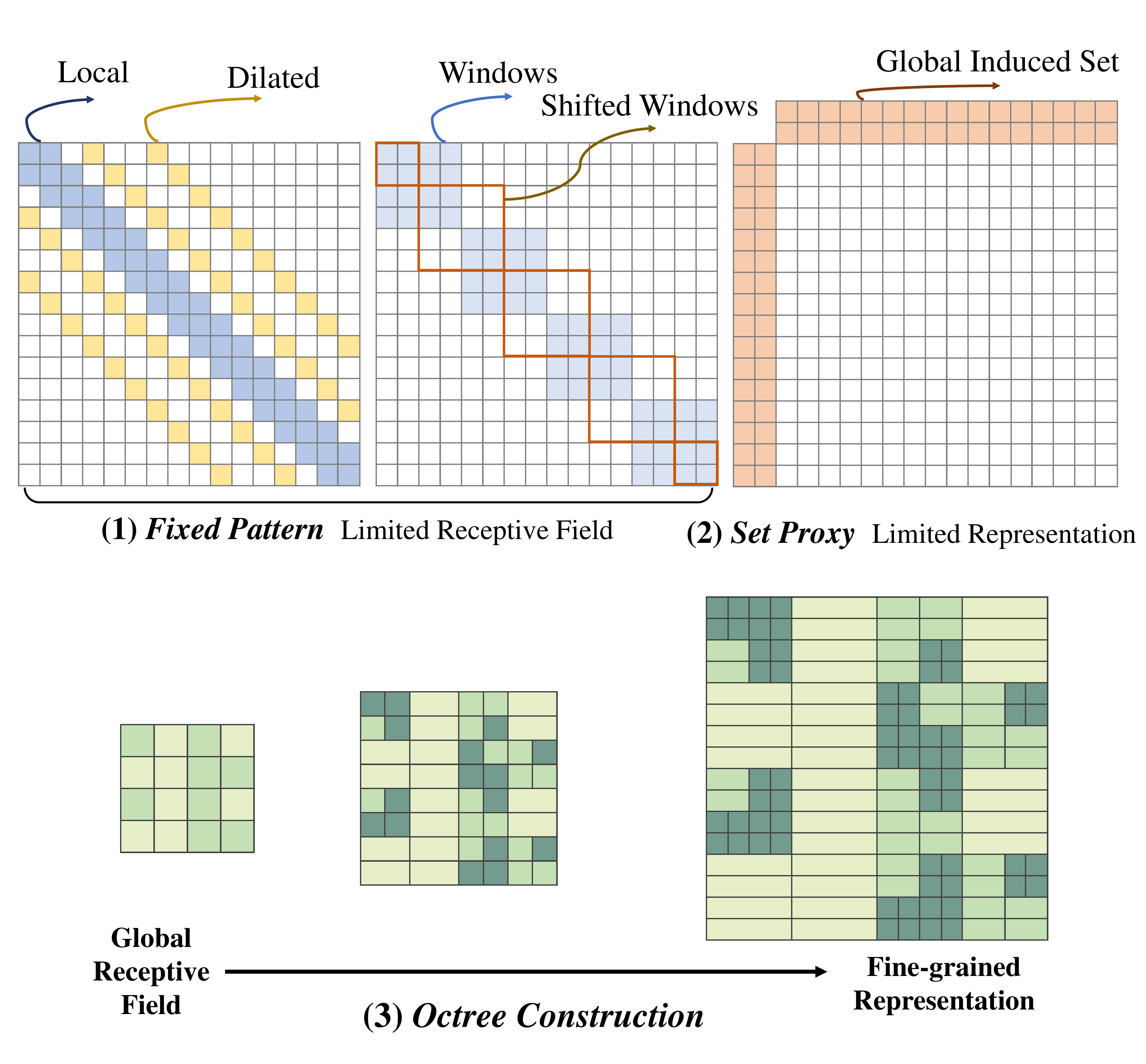}

    \caption{Illustration of three sparsification strategies of attention matrices. Fixed pattern (1) narrows receptive fields and set proxy (2) discards elaborate correlations. The proposed octree construction (3) keeps the global receptive field in a coarse-grained manner while maintaining fine-grained representations.}
    \label{motivation}
\end{figure}

3D object detection from point clouds has received extensive attention during the past decade for its ability to provide accurate and stable recognition and localization in autonomous driving perception systems. In this task, feature learning plays a very fundamental and crucial role; yet it is rather challenging due to not only the disordered and sparse nature of data sampling, but also to insufficient acquisition under occlusion or at a distance. To address this issue, many methods have been proposed, which can be taxonomized into two major classes, \emph{i.e.} grid-based and point-based. The former first regularize point clouds into multi-view images or voxels and then apply 2D or 3D CNNs to build shape representations~\cite{yan2018second, chen2022focal}, while the latter directly conduct MLP based networks such as PointNet++~\cite{qi2017pointnet++} and DGCNN~\cite{wang2019dynamic} on original points for geometry description~\cite{shi2019pointrcnn, qi2018frustum, zhang2021pc, shi2020point}. Unfortunately, they fail to capture necessary context information through the small receptive fields in the deep models, leading to limited results.

Witnessing the recent success of Transformers in NLP, many studies have investigated and extended such architectures for 3D vision~\cite{zhao2021point, pan20213d, mao2021voxel, zhang2022cat}. Transformers are reputed to model long-range dependencies, delivering global receptive fields, and to be suitable for scattered inputs of arbitrary sizes. Meanwhile, in contrast to those static weights that are learned in convolutions, Transformers dynamically aggregate the input features according to the relationships between tokens. Regarding the case in 3D object detection, compared to point-based Transformers~\cite{zhao2021point, guo2021pct}, voxel-based ones show the superiority in efficiency. However, they tend to suffer heavy computations when dealing with large scale scenes because of the quadratic complexity of Transformers, with the underlying dilemma between the grid size and the grid amount in voxelization. Taking the KITTI dataset as an example, it is unrealistic for Transformers to operate on the feature map with the spatial shape of $200 \times 176 \times 5$, which is commonly adopted in most of the detection heads\cite{yan2018second, shi2020pv, openpcdet2020, yin2021center}.

More recently, there have appeared an influx of efficient self-attention model variants that attempt to tackle long sequences as input. They generally sparsify the attention matrix by fixed patterns~\cite{qiu2019blockwise, liu2021swin, dong2022cswin}, learned patterns~\cite{liang2022not, tang2022quadtree} or a combination of them~\cite{ainslie2020etc, zaheer2020big}. Fixed patterns chunk the input sequence into blocks of local windows or dilation windows, whilst learned patterns determine a notion of token relevance and eliminate or cluster outliers. Specific to 3D object detection from point clouds, VoTr~\cite{mao2021voxel} modifies self-attention with pre-defined patterns including local windows and stride dilation ones in a sparse query manner, and the dilation mechanism enlarges the receptive field by sampling attending tokens in a radius. SST~\cite{fan2022embracing} splits input tokens into non-overlapping patterns in a block-wise way and enables window shifting to capture cross-window correlation. Despite some improvements reported, they both only achieve bigger local receptive fields rather than the expected global ones, and computations still increase rapidly with the expansion of receptive fields.

Another alternative on self-attention is to take advantage of a proxy memory bank which has the access to the entire sequence tokens~\cite{ainslie2020etc, zaheer2020big, beltagy2020longformer}. By using a small number of induced proxies to compress the whole sequence, it diffuses the global context efficiently. VoxSet~\cite{he2022voxel} adapts Set Transformer\cite{lee2019set} to 3D object detection and exploits an induced set to model a set-to-set point cloud translation. With the help of the compressed global proxies and Conv-FFN, it obtains a global receptive field; nevertheless, as they admit, it is sub-optimal to set only a few latent codes as proxies for a large 3D scene, prone to impairing the representation of different point cloud structures and their correlations. Therefore, there remains space for a stronger solution.

In this paper, we present a novel Transformer network, namely \textbf{Oc}tree-based \textbf{Tr}ansformer (\textbf{OcTr}), for 3D object detection. We firstly devise an octree-based learnable sparse pattern, \emph{i.e.} \textit{OctAttn}, which meticulously and efficiently encodes point clouds of scenes as shown in Fig.~\ref{motivation}. The \textit{OctAttn} module constructs a feature pyramid by gathering and applies self-attention to the top level of the feature pyramid to select the most relevant tokens, which are deemed as the octants to be divided in the subsequent. When propagating to the level below, the key/value inputs are restricted by the octants from the top. Through recursively conducting this process, \textit{OctAttn} captures rich global context features by a global receptive field in a coarse-to-fine manner while reducing the quadratic complexity of vanilla self-attention to the linear complexity. In addition, for better foreground perception, we propose a hybrid positional embedding, which consists of the semantic-aware positional embedding and attention mask, to fully exploit geometry and semantic clues. Thanks to the designs above, OcTr delivers a competitive trade-off between accuracy and efficiency.

Our contribution is summarized in three-fold:
\begin{enumerate}
    \item We propose OcTr for voxel-based 3D object detection, which efficiently learns enhanced representations by a global receptive field with rich contexts.
    \item We propose an octree-based learnable attention sparsification scheme (\textit{OctAttn}) and a hybrid positional embedding combining geometry and semantics. 
    \item We carry out experiments on the Waymo Open Dataset (WOD) and the KITTI dataset and report state-of-the-art performance with significant gains on far objects.
\end{enumerate}

\begin{figure*}[t]
    \centering
    \includegraphics[width=1.0\textwidth]{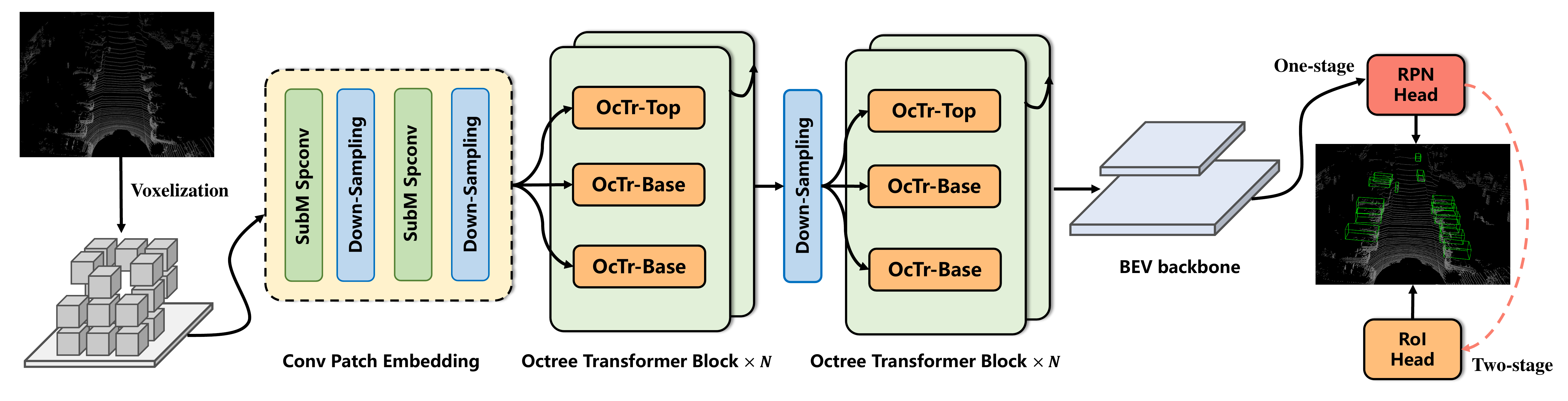}
    \caption{Framework overview of the proposed Octree-based Transformer (OcTr) model. }
    \label{pipeline}
\end{figure*} 

\section{Related Work}
\label{sec:formatting}

\subsection{3D Object Detection from Point Clouds}

There exist two prevailing point-cloud representations in 3D object detection, \emph{i.e.} point-based and voxel-based.

The point-based methods~\cite{shi2020point, shi2019pointrcnn, qi2019deep, zhang2021pc} directly process raw point clouds in the irregular 3D space. As a pioneering attempt, F-Pointnet \cite{qi2018frustum} employs instance segmentation in frustums to extract proposals. VoteNet \cite{qi2019deep} clusters objects from the surface in a deep Hough voting manner. PointRCNN \cite{shi2019pointrcnn} generates 3D RoIs with foreground segmentation and applies an RCNN-style \cite{ren2015faster} two-stage refinement. Different from PointRCNN, some existing methods \cite{yang20203dssd, zhang2022not, chen2022sasa} build a lightweight and efficient single stage 3D object detection framework. However, the current point-based methods still suffer from a large computation burden, which is not suitable for large-scale point cloud scenes.

The voxel-based ones \cite{zhou2018voxelnet, yan2018second, zheng2021cia, he2020structure, deng2021voxel, yin2021center, chen2022focal} conduct voxelization on entire point clouds to construct regular grids. VoxelNet \cite{zhou2018voxelnet} exploits the voxel feature encoding layer with 3D convolutions to extract the feature of each voxel. SECOND \cite{yan2018second} improves the model with sparse 3D convolutions, significantly increasing the speed of both training and inference. Pointpillars \cite{lang2019pointpillars} compacts point clouds into vertical columns and encodes them with 2D CNNs. Several recent methods \cite{shi2020pv, he2020structure, yang2019std} also explore merging point-based and voxel-based networks into one framework for complementary features from different representations of point clouds. Unfortunately, they all use small convolution kernels with limited receptive fields, which are not competent to capture global context that is important to 3D detection.

\subsection{Transformer in 3D Vision}

Inspired by the great success of the self-attention mechanism in NLP \cite{vaswani2017attention} and CV \cite{dosovitskiy2020image, liu2021swin}, Transformers have been adapted to 3D vision for their ability to capture long-range dependencies. For instance, Point Transformer \cite{zhao2021point} brings in vector attention that modulates individual feature channels for point cloud classification and segmentation; PCT \cite{guo2021pct} presents offset-attention with the implicit Laplace operator and normalization refinement which is more suitable for point cloud learning. To address the high latency, some methods \cite{park2022fast, zhang2022patchformer} adopt voxels or patches for acceleration.

For 3D detection, 3DETR \cite{misra2021end} treats and predicts bounding boxes as sequences in an end-to-end manner. CT3D \cite{sheng2021improving} leverages a channel-wise Transformer architecture to refine the RoI head. To learn context-aware representations, some studies \cite{pan20213d, mao2021voxel, he2022voxel, fan2022embracing} introduce Transformers into a point- or voxel-based encoder. Pointformer \cite{pan20213d} stacks local, global and local-global Transformers based on the point-based encoder; VoTr \cite{mao2021voxel} exploits dilated attention with fast query to enlarge receptive fields; VoxSet \cite{he2022voxel} builds an induced point set as proxies of global context and applies point-to-point translation using voxels as mediums; and SST \cite{fan2022embracing} embraces single strides without down-sampling and conducts window attention combining with its shifted version. Even though they all expand receptive fields by diverse Transformer variants, global context is not adequately involved or efficiently utilized. In contrast, we propose Octree-based Transformer (OcTr) for voxel-based 3D object detection, achieving a true global receptive field that balances accuracy and efficiency. 

\section{Method}

\subsection{Framework}

\begin{figure*}[t]
    \centering
    \includegraphics[width=1.0\textwidth]{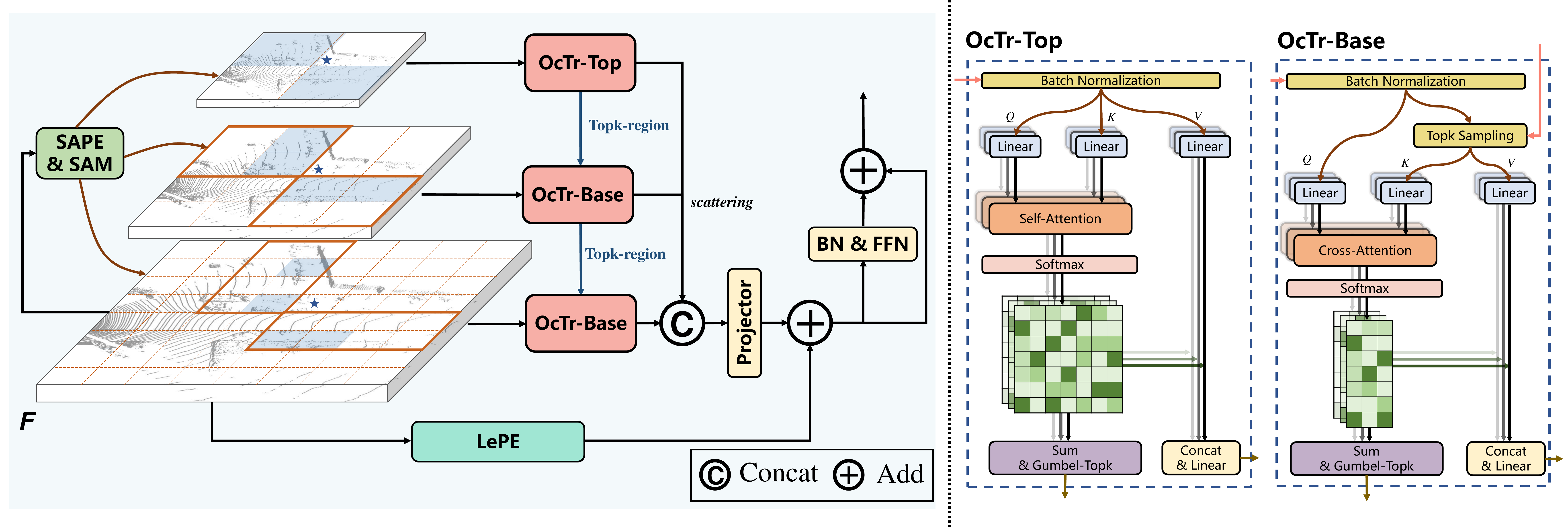}
    \caption{\textbf{Octree Attention} (\textit{OctAttn}). The \textit{OctAttn} module constructs a dynamic octree on a hierarchical feature pyramid with a partitive criterion of attention scores. For illustration, an octree representation is individually pruned from the pyramid for each grid in $F$. For example, the blue grids are the divided octants for the grids with stars. The detailed structures of \textbf{OcTr-Top} and \textbf{OcTr-Base} are shown at the right side.
    }
    \label{octattn}
    \vspace{-1.2em}
\end{figure*}

This sub-section describes the overall framework of the proposed OcTr model as shown in Fig.~\ref{pipeline}. Specifically, as inspired by \cite{xiao2021early}, we first voxelize the point cloud into regular grids and adopt the sparse 3D convolution for patch embedding, where the girds are regarded as the ``tokens'' and are passed through the Octree Transformer Blocks (OTB). Compared with the vanilla Transformer block, the self-attention module is substituted by the proposed octree-attention \textit{OctAttn}, which encodes global context in a more efficient way. After applying a hybrid semantic embedding on multi-scale features, we sequentially stack two OTBs, tailed by a down-sampling layer. The voxel features are then projected into the BEV view by point-wise convolutions and are passed through a multi-scale dense 2D backbone. Ultimately, an anchor-based or anchor-free RPN head is used for 3D proposal generation, and the RoI head is optional for refinement. Note that our OcTr can be adopted for most of the voxel-based detection frameworks by simply altering 3D backbones.

\subsection{Self-attention Revisit}

According to \cite{vaswani2017attention}, Transformer encoder blocks typically include a multi-head self-attention (MHSA) mechanism, a feed-forward network (FFN), a normalization function, and the residual connections \cite{he2016deep}. Given an input sequence $X$, the principle part of MHSA is formulated as
\begin{equation}
    \label{eq:transformer}
    \begin{aligned}
    \text{MHSA}(\it{X}) = \sum_{h=1}^{H} W_{h}[\sigma(\frac{\it{X}W_{q}W_{k}^{T}\it{X}^{T}}{\sqrt{d}})\cdot\it{X}W_{v}],
    \end{aligned}
\end{equation}
\noindent where $h$ denotes the index of the head, and $H$, $\sigma$, $W$ and $d$ are the amount of heads, softmax function, learnable weight and feature dimension, respectively. The subscripts of $q$, $k$ and $v$ indicate query, key and value. The inputs and outputs of the MHSA module are connected by residual connectors and normalization layers. The MLP-based FFN connects its inputs/outputs in a similar manner. 

\subsection{Octree Attention}

An octree is a multi-scale asymmetric and efficient representation for unstructured 3D data such as point clouds. To build an octree for the input point cloud, we recursively sub-divide it in the breadth-first order until the pre-defined octree depth is reached. Whether to sub-divide an octant is determined by the occupancy \cite{wang2017cnn, riegler2017octnet}, the surface approximation \cite{tang2021octfield, wang2018adaptive} or a learning algorithm \cite{martel2021acorn}. 

Despite of being resource-friendly, the octree representation is discrete and non-differentiable, which motivates us to ameliorate previous cumbersome pre-processing division by introducing a novel octree-based attention mechanism.As depicted in \cite{liang2022not, tang2022quadtree}, the attention matrices calculated by self-attention imply the relevance of input tokens and guide feature selection. We thus prune the dense attention matrices of the multi-scale feature pyramid to sparse octree attention in an adaptive and parallel manner, namely \textit{OctAttn}.

As shown in Fig.~\ref{octattn}, let the output feature map and coordinates of convolutional patch embedding be $F_{0} \in \mathbb{R}^{M \times d}$ and $I_{0} \in \mathbb{R}^{M \times 3}$; $M$ and $d$ indicate the amount of the non-empty grids in a batch and feature dimension, respectively. Based on $F_{0}$, we generate a multi-scale feature pyramid as 

\begin{equation}
\begin{aligned} C=\{F_{n}, I_{n}\}^{N}, n \in [0, N),
\end{aligned}
\end{equation}

\noindent where

\begin{equation}
    \label{downsample}
    \begin{aligned} I_{n} = \lfloor \frac{I_{0}}{2^{n}} \rfloor, \ \ F_{n} = \text{BN}(\mathcal{S}_{max}(F_{0}, I_{n})),
    \end{aligned}
\end{equation}

\noindent $n$, $N$, BN and $\mathcal{S}_{max}$ are the index of the level of the multi-scale feature pyramid, the height of the pyramid, batch normalization~\cite{ioffe2015batch} and the max scatter function, respectively.

The pruning begins from the top of the pyramid. The top feature map, \emph{i.e.} $F_{N-1}$, is reorganized to dense input tokens $\bar{F}_{N-1} \in \mathbb{R}^{B \times m_{N-1} \times d}$, where $B$ and $m_{N-1}$ are the batch size and maximum number of non-empty voxels per batch, respectively. The voxel is padded with 0 if it is empty. As shown in Eq. \eqref{selfattn}, the MHSA takes $\bar{F}_{N-1}$ as input and outputs the attention score matrices $\mathcal{A}_{N-1} \in \mathbb{R}^{B \times m_{N-1} \times m_{N-1}}$ and attentive features $\bar{F}_{N-1}^{'}$ as

\begin{equation}
    \label{selfattn}
    \vspace{0.4em}
    \begin{aligned}
    & \mathcal{A}_{N-1} = \sum_{h=1}^{H} \sigma(\frac{\it{\bar{F}_{N-1}}W_{q}W_{k}^{T}\it{\bar{F}_{N-1}}^{T}}{\sqrt{d}}),
    \\
    & \bar{F}_{N-1}^{'} = \sum_{h=1}^{H} W_{h}[\sigma(\frac{\it{\bar{F}_{N-1}}W_{q}W_{k}^{T}\it{\bar{F}_{N-1}}^{T}}{\sqrt{d}})\cdot\it{\bar{F}_{N-1}}W_{v}].
    \end{aligned}
\end{equation}

\noindent For each query token, we select the top$k$ attention scores as its most relevant token group in a row-wise way, denoted by $O_{N-1} \in \mathbb{Z}^{B \times m_{N-1} \times k}$. 

When propagating from top to bottom through the pyramid and reaching the $n$-th level, we uniformly sample limited $K$ attending octants from the selected regions with features in $\bar{F}_{n}$ and top$k$ indices in $O_{n+1}$. We conduct cross-attention instead of self-attention with the dense query sequence of the $n$-th level, $\bar{Q}_{n} \in \mathbb{R}^{B \times m_{n} \times d} $, and the compact sampled key/value sequence of the $n$-th level, $\bar{K}_{n}/\bar{V}_{n} \in \mathbb{R}^{B \times m_{n} \times K \times d}$ , which is formulated as below

 \begin{equation}
    \label{crossattn}
    \begin{aligned}
    & \mathcal{A}_{n} = \sum_{h=1}^{H} \sigma(\frac{\it{\bar{Q}_{n}}W_{q}W_{k}^{T}\it{\bar{K}_{n}}^{T}}{\sqrt{d}}),
    \\
    & \bar{F}_{n}^{'} = \sum_{h=1}^{H} W_{h}[\sigma(\frac{\it{\bar{Q}_{n}}W_{q}W_{k}^{T}\it{\bar{K}_{n}}^{T}}{\sqrt{d}})\cdot\it{\bar{V}_{n}}W_{v}].
    \end{aligned}
\end{equation}

\noindent Backing off the sampling, this can be treated as an attention mask on self-attention matrices. The above process is run recursively until reaching the bottom level of the pyramid.

Furthermore, as the top$k$ selection is a hard decision that disables gradient back-propagation, we adopt the Gumbel-top$k$ technique~\cite{jang2016categorical} to perform a differentiable and continuous approximation by replacing the vanilla top$k$ selection. The normalized scores used in top$k$ are derived from the distribution in Eq. \eqref{gumbel} during training, maintaining the original ones during inference. $g$, $\tau$, $m_{n}$ denote the noise sampled from the Gumbel distribution, the temperature and the amount of non-empty voxels in layer $n$, respectively.

\begin{equation}
    \label{gumbel}
    \begin{aligned}
    & p_{i} = \frac{\text{exp}((\mathcal{A}_{n}^{i}+g_{i})/\tau)}{\sum \limits _{i}^{m_{n}} \text{exp}((\mathcal{A}_{n}^{i}+g_{i})/\tau)} \ \in [0, 1].
    \end{aligned}
\end{equation}

In order to leverage the multi-scale features in distinct spatial shapes, we concatenate them by upsampling, which is implemented by inverse indices of the scatter function, followed by a linear projection layer for aligning the input feature dimension.

As the local context is generally critical for object detection, inspired by \cite{dong2022cswin}, we additionally introduce a Locally enhanced Positional Embedding (LePE) which enables local neighbor interactions on the value sequence. With sub-manifold sparse convolutions, we replace the residual connections in the attention mechanism with LePE.

Finally, OTB is formulated as follows

\begin{equation}
    \label{octr}
    \begin{aligned}
    & \tilde{F} \ = \ \text{FC}(\{F_{N-1}^{'}|F_{N-2}^{'}|...|F_{0}^{'}\}) + \text{LePE}(F_{0}),
    \\
    & \tilde{F}^{'} \ = \ \text{BN}(\text{FFN}(\tilde{F})) + \tilde{F},
    \end{aligned}
\end{equation}

\noindent where $F_{n}^{'} \in \mathbb{R}^{m_{0} \times d}$ denotes the compact tensor of the upsampled $\bar{F}_{n}^{'}$, $\vert$ indicates concatenation, and FC denotes the fully-connected layer.

Besides, we analyze the time complexity of the octree attention as below
\begin{equation}
    \label{timecomp}
    \begin{aligned}
    & \mathcal{O}((\frac{M}{\omega^{N-1}})^{2}+\sum \limits_{n=0}^{N-2} \frac{KM}{\omega^{n}}) \\
    = & \mathcal{O}((\frac{M}{\omega^{N-1}})^{2}+ \frac{\omega}{\omega-1}KM(1-\omega^{1-N})),
    \end{aligned}
\end{equation}
where $\omega$ is the average down-sampling ratio in the sparse voxel representation.
\subsection{Semantic Positional Embedding}

\begin{figure}[t]
    \centering
    \includegraphics[width=0.47\textwidth]{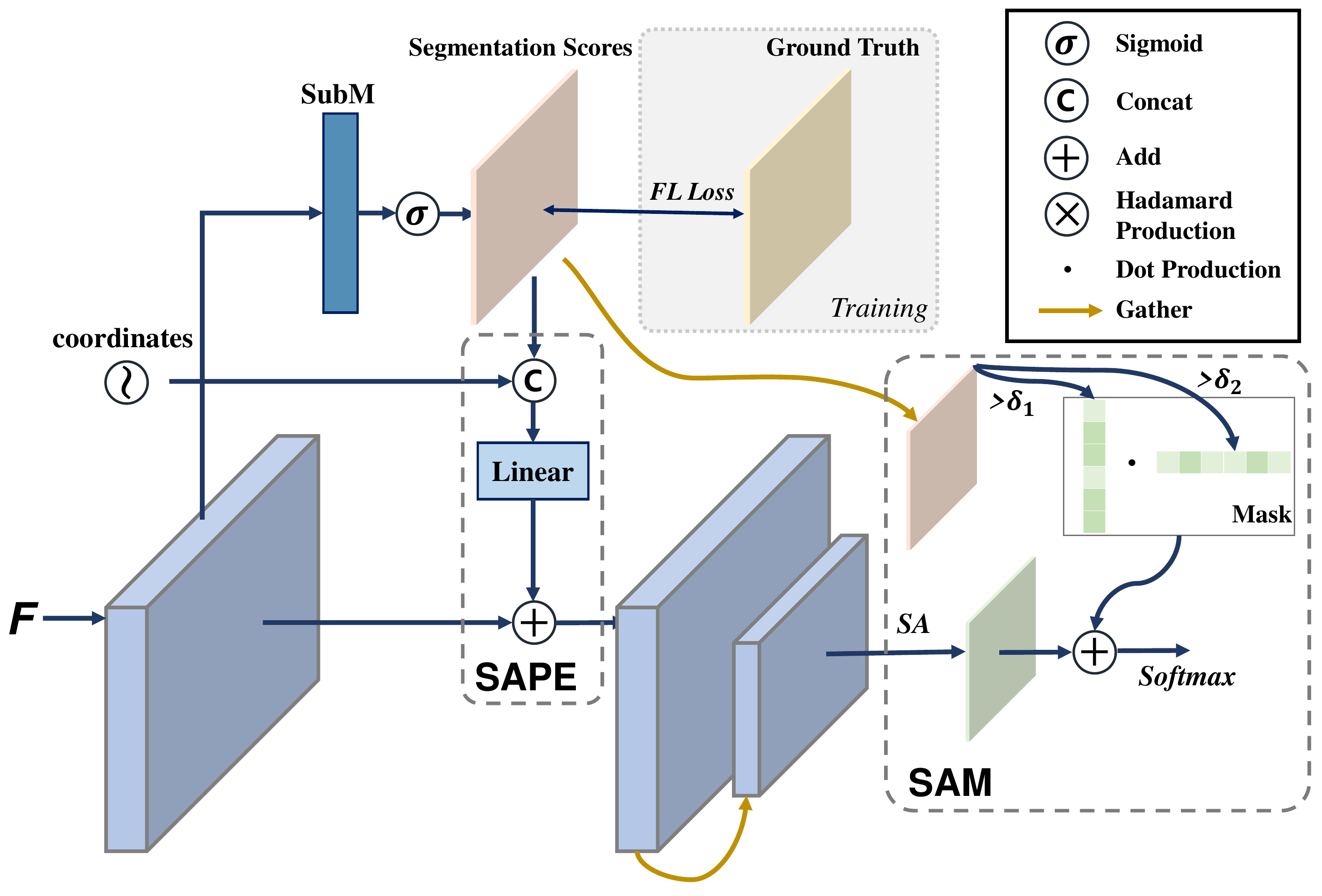}
    \caption{Illustration of semantic positional embedding, where an extra foreground segmentation branch is adopted with supervision and semantic scores are concatenated as absolute positional embeddings and serve as relative masks on attention matrices.}
    \label{fig_pe}
    \vspace{-0.5cm}
\end{figure}

Due to the large proportion of background grids in point clouds, the attention matrices are dominated by background grid pairs, leading to a sub-optimal solution. To fully leverage the local 3D shape patterns and original voxel coordinates, we propose a hybrid positional embedding to capture both the geometry and semantic clues as displayed in Fig.~\ref{fig_pe}.  

Specifically, we first segment foreground grids using the supervision from the ground truth. Segmentation scores are predicted by a sub-manifold sparse convolution branch with a sigmoid function; and the focal loss \cite{lin2017focal} is applied to balance the foreground and background.  

We concatenate the center coordinate ($x$, $y$, $z$) and the semantic $score$ with the feature $f$ in a grid-wise manner, followed by a linear projection without the bias as below
\begin{equation}
    \label{ape}
    \begin{aligned}
    \text{SAPE}(X) & = \text{FC}_{d+4 \rightarrow d}(\{x,y,z,score | f\}) \\
    &= \text{FC}_{d \rightarrow d}(f) + \text{FC}_{4 \rightarrow d}(x,y,z,score).
    \end{aligned}
\end{equation}
Eq. \eqref{ape} is equivalent to the absolute positional embedding (APE), thus being denoted as the Semantic APE (SAPE). The scatter function such as the mean, max and batch normalization naturally provide position and semantic information of the downsampled voxel grids, making it applicable for the multi-scale feature pyramid.

Besides the semantic clues implicitly used in SAPE, we employ the Semantic Attention Mask (SAM) based on the segmentation scores and we mask the attention matrices to address correlations between inferior queries and superior keys in a simple yet effective way. Given a scalar attention matrix $\mathcal{A} \in \mathbb{R}^{N_{q} \times N_{k} }$ before softmax and the segmentation scores of query and key/value $S_{q} \in \mathbb{R}^{N_{q}}$ and $S_{k} \in \mathbb{R}^{N_{k}}$, we formulate the attention matrix after softmax as
\begin{equation}
    \label{rpe}
    \begin{aligned}
    \mathcal{A}^{'} = \sigma(-\Gamma \cdot [1-(S_{q} \geq \delta_{q})(S_{k} \geq \delta_{k})^{T}] + \mathcal{A}), \\
    \end{aligned}
    \vspace{-0.3em}
\end{equation}
where $N_{q}$, $\delta_{q}$, $N_{k}$, $\delta_{k}$, $\sigma$ and $\Gamma$ are the length and threshold of the query sequence, length and threshold of the key sequence, softmax function and infinite scalar, respectively. 

Finally, we broadcast the semantic mask $\mathcal{A}^{'}$ to all heads.

\begin{table*}[t]
    \centering
    \begin{tabular}{l|c|c|c|c|c|c}\toprule[1pt]
    \multicolumn{1}{l|}{\multirow{2}*{\textbf{Model}}} &\multicolumn{1}{c|}{\textbf{Vehicle (L1)}} & \multicolumn{1}{c|}{\textbf{Vehicle (L2)}} &\multicolumn{1}{c|}{\textbf{Pedes. (L1)}} & \multicolumn{1}{c|}{\textbf{Pedes. (L2)}} &\multicolumn{1}{c|}{\textbf{Cyclist (L1)}} & \multicolumn{1}{c}{\textbf{Cyclist (L2)}} \\ 
    ~ & \textbf{mAP/mAPH} & \textbf{mAP/mAPH} & \textbf{mAP/mAPH} & \textbf{mAP/mAPH} & \textbf{mAP/mAPH}  & \textbf{mAP/mAPH} \\ \hline
    SECOND\cite{yan2018second} & 70.96/70.34 & 62.58/62.02 & 65.23/54.24 & 57.22/47.49 & 57.13/55.62 & 54.97/53.53 \\
    PointPillar\cite{lang2019pointpillars} & 70.43/69.83 & 62.18/61.64 & 66.21/46.32 & 58.18/40.64 & 55.26/51.75 & 53.18/49.80 \\
    PartA$^{2}$Net\cite{shi2020points} & 74.82/74.32 & 65.88/65.42 & 71.76/63.64 & 62.53/55.30 & 67.35/66.15 & 65.05/63.89 \\
    PVRCNN\cite{shi2020pv} & 75.41/74.74 & 67.44/66.80 & 71.98/61.24 & 63.70/53.95 & 65.88/64.25 & 63.39/61.82 \\
    CenterPoint\cite{yin2021center} & 71.33/70.76 & 63.16/62.65 & 72.09/65.49 & 64.27/58.23 & 68.68/67.39 & 66.11/64.87 \\
    LiDAR-RCNN\cite{li2021lidar} & 73.5/73.0 & 64.7/64.2 & 71.2/58.7 & 63.1/51.7 & 68.6/66.9 & 66.1/64.4 \\ 
    Voxel-RCNN\cite{deng2021voxel} & 75.59/- & 66.59/- & -/- & -/- & -/- & -/- \\
    PVRCNN++\cite{shi2021pv} & 77.82/77.32 & 69.07/68.62 & 77.99/71.36 & 69.92/63.74 & 71.80/70.71 & 69.31/68.26 \\
    SST$^{\dag}$\cite{fan2022embracing} & 76.22/75.79 & 68.04/67.64 & 81.39/74.05 & 72.82/65.93 & -/- & -/- \\
    PDV\cite{hu2022point}  & 76.85/76.33 & 69.30/68.81 & 74.19/65.96 & 65.85/58.28 & 68.71/67.55 & 66.49/65.36 \\
    \hline
    Ours & \textbf{78.12/77.63} & \textbf{69.79/69.34} & \textbf{80.76/74.39} & \textbf{72.48/66.52} & \textbf{72.58/71.50} & \textbf{69.93/68.90} \\
    \bottomrule
    \end{tabular}
    \caption{Performance on WOD with 202 \textit{validation} sequences for vehicle (IoU=0.7), pedestrian (IoU=0.5) and cyclist (IoU=0.5), using 20\% samples for training. All the results are achieved by the models simultaneously trained for 3 classes on single frames, except the ones of the model marked by $\dag$, which is only trained for a single class. Refer to \textit{Supp. B} for the results trained with 100\% samples.}
    \label{waymoval}
    \vspace{-0.4cm}
\end{table*}

\begin{table}[h!]
    \centering
    \resizebox{0.47\textwidth}{!}{
    \begin{tabular}{l|cccc}\toprule[1pt]
    \multicolumn{1}{l|}{\multirow{2}*{\textbf{Model}}} &\multicolumn{4}{c}{\textbf{mAP$_{3D}$ (L1)@Vehicle}} \\
    ~ & \textbf{Overall} & \textbf{0-30m} & \textbf{30m-50m} & \textbf{50m-inf} \\ \midrule
    PV-RCNN\cite{shi2020pv} & 70.30 & 91.92 & 69.21 & 42.17 \\
    Voxel-RCNN\cite{deng2021voxel} & 75.59 & 92.49 & 74.09 & 53.15 \\
    VoTR-TSD\cite{mao2021voxel} & 74.95 & 92.28 & 73.36 & 51.09 \\
    CT3D\cite{sheng2021improving} & 76.30 & 92.51 & 75.07 & 55.36 \\
    Pyramid\_PV\cite{mao2021pyramid} & 76.30 & 92.67 & 74.91 & 54.54 \\
    PDV\cite{hu2022point} & 76.85 & \textbf{93.13} & 75.49 & 54.75 \\
    VoxSeT\cite{he2022voxel} & 77.82 & 92.78 & 77.21 & 54.41 \\ 
    \hline
    Ours & \textbf{78.82} & 92.99 & \textbf{77.66} & \textbf{58.02} \\ \hline \hline
    \multicolumn{1}{l|}{\multirow{2}*{\textbf{Model}}} &\multicolumn{4}{c}{\textbf{mAP$_{3D}$ (L2)@Vehicle}} \\
    ~  & \textbf{Overall}  & \textbf{0-30m} & \textbf{30-50m} & \textbf{50m-inf} \\ \hline
    PV-RCNN\cite{shi2020pv} & 65.36 & 91.58 & 65.13 & 36.46 \\
    Voxel-RCNN\cite{deng2021voxel} & 66.59 & 91.74 & 67.89 & 40.80 \\
    CT3D\cite{sheng2021improving} & 69.04 & 91.76 & 68.93 & 42.60 \\
    PDV\cite{hu2022point} & 69.30 & \textbf{92.41} & 69.36 & 42.16 \\
    VoxSeT\cite{he2022voxel} & 70.21 & 92.05 & 70.10 & 43.20 \\ 
    \hline
    Ours & \textbf{70.50} & 91.78 & \textbf{71.28} & \textbf{45.46} \\
    \bottomrule
    \end{tabular}
    }
    \caption{Results on the WOD \textit{validation} set in different ranges for vehicle detection.}
    \label{waymorange}
    \vspace{-0.5cm}
\end{table}

\section{Experiments}

We evaluate the proposed OcTr network on the Waymo Open Dataset (WOD) and KITTI dataset, both of which are popular in 3D object detection. In this section, we introduce the benchmarks and implementation details, make comparison to the previous state-of-the-art counterparts, and ablate the key designs of OcTr.

\subsection{Datasets and Implementation Details}

\begin{table*}[h!]
    \centering
    \resizebox{0.75\textwidth}{!}{
    \begin{tabular}{l|ccc|ccc|ccc}\toprule[1pt]
    \multicolumn{1}{l|}{\multirow{2}*{\textbf{Model}}} &\multicolumn{3}{c|}{\textbf{mAP$_{3D}$@Car}} &\multicolumn{3}{c|}{\textbf{mAP$_{3D}$@Pedestrian}} &\multicolumn{3}{c}{\textbf{mAP$_{3D}$@Cyclist}}\\ 
    ~  & \textbf{Easy} & \textbf{Mod.} & \textbf{Hard} & \textbf{Easy} & \textbf{Mod.} & \textbf{Hard} & \textbf{Easy} & \textbf{Mod.} & \textbf{Hard} \\ \hline
    SECOND\cite{yan2018second} & \textbf{88.61} & \textbf{78.62} & \textbf{77.22} & 56.55 & 52.98 & 47.73 & 80.58 & 67.15 & 63.10  \\ 
    PointPillars\cite{lang2019pointpillars} & 88.46 & 77.28 & 74.65 & 57.75 & 52.29 & 47.90 & 80.04 & 62.61 & 59.52  \\
    VoTR\cite{mao2021voxel} & 87.86 & 78.27 & 76.93 & - & - & - & - & - & -  \\ 
    VoxSeT\cite{he2022voxel} & 88.45 & 78.48 & 77.07 & 60.62 & 54.74 & 50.39 & 84.07 & 68.11 & 65.14  \\ \hline
    Ours & 88.43 & 78.57 & 77.16 & \textbf{61.49} & \textbf{57.17} & \textbf{52.35} & \textbf{85.29} & \textbf{70.44} & \textbf{66.17}  \\
    \bottomrule
    \end{tabular}
    }
    \caption{Results of the single-stage models on the KITTI \textit{val} set. All the models adopt the same anchor-based region proposal network as the detection head. ``Mod.'' denotes the moderate difficulty level.}
    \label{kittisingle}
\end{table*}

\begin{table*}[h!]
    \centering
    \resizebox{0.7\textwidth}{!}{
    \begin{tabular}{l|cccc|cccc}\toprule[1pt]
    \multicolumn{1}{l|}{\multirow{2}*{\textbf{Model}}} &\multicolumn{4}{c|}{\textbf{mAP$_{3D}$@Car on test}} &\multicolumn{4}{c}{\textbf{mAP$_{3D}$@Car on val}}\\
    ~ & \textbf{Easy} & \textbf{Mod.} & \textbf{Hard} & \textbf{Mean} & \textbf{Easy} & \textbf{Mod.} & \textbf{Hard} & \textbf{Mean}\\ \midrule
    SECOND\cite{yan2018second} & 83.34 & 72.55 & 65.82 & 73.90 & 88.61 & 78.62 & 77.22 & 81.48 \\ 
    PointPillars\cite{lang2019pointpillars} & 82.58 & 74.31 & 68.99 & 75.29 & 86.62 & 76.06 & 68.91 & 77.20 \\
    STD\cite{yang2019std} & 87.95 & 79.71 & 75.09 & 80.92 & 89.70 & 79.80 & \textbf{79.30} & 82.93 \\
    SA-SSD\cite{he2020structure} & 88.75 & 79.79 & 74.16 & 80.90 & \textbf{90.15} & 79.91 & 78.78 & 82.95 \\ 
    3DSSD\cite{yang20203dssd} & 88.36 & 79.57 & 74.55 & 80.83 & 89.71 & 79.45 & 78.67 & 82.61 \\
    PV-RCNN\cite{shi2020pv} & 90.25 & 81.43 & 76.82 & 82.83 & 89.35 & 83.69 & 78.70 & 83.91 \\ 
    Voxel-RCNN\cite{deng2021voxel} & \textbf{90.90} & 81.62 & 77.06 & 83.19 & 89.41 & 84.52 & 78.93 & 84.29 \\
    CT3D\cite{sheng2021improving} & 87.83 & 81.77 & 77.16 & 82.25 & 89.54 & 86.06 & 78.99 & \underline{84.86} \\ 
    VoTR-TSD\cite{mao2021voxel} & 89.90 & 82.09 & \textbf{79.14} & \underline{83.71} & 89.04 & 84.04 & 78.68 & 83.92 \\
    VoxSeT\cite{he2022voxel} & 88.53 & 82.06 & 77.46 & 82.68 & 89.21 & \underline{86.71} & 78.56 & 84.83 \\ 
    Focals Conv\cite{chen2022focal} & 90.55 & \underline{82.28} & 77.59 & 83.47 & 89.52 & 84.93 & 79.18 & 84.54 \\
    \hline
    Ours & \underline{90.88} & \textbf{82.64} & \underline{77.77} & \textbf{83.76} & \underline{89.80} & \textbf{86.97} & \underline{79.28} & \textbf{85.35} \\
    \bottomrule
    \end{tabular}
    }
    \caption{Comparison to the state-of-the-art models on the KITTI \textit{test} and \textit{val} sets. ``Mod.'' and  ``Mean'' denote the moderate difficulty level and the average mAP for the three levels, respectively. The best results are \textbf{bolded} and the second best ones are \underline{underlined}.}
    \label{kittitwotest}
    \vspace{-0.2cm}
\end{table*}

\noindent \textbf{WOD} \cite{sun2020scalability} is a large dataset of autonomous driving scenes. It totally contains 798 training sequences with around 160K LiDAR samples and 202 validation sequences with 40K LiDAR samples, with the mean Average Precision (mAP) and mAP weighted by heading accuracy (mAPH) as evaluation metrics. There are also two levels of difficulty describing the sparsity in each bounding box, and LEVEL\_1 (L1) and LEVEL\_2 (L2) denote more than 5 points and 1-5 points, respectively. For detection performance along distance, it provides mAP/mAPH on 0-30m, 30m-50m and 50m-inf.

\noindent \textbf{KITTI} \cite{geiger2012we} is a widely used benchmark for 3D object detection, which includes 3,712, 3,769 and 7,518 frames for training, validation and testing, respectively. mAP is used as the official metric with 11 recall points for the \textit{val} set and 40 for the \textit{test} set, and the IoU thresholds are set to 0.7, 0.5, and 0.5 for car, pedestrian and cyclist. We use the official setting in all experiments.  

\noindent \textbf{Implementation Details} The total loss for optimizing the overall two-stage detection is formulated as Eq.~\eqref{eq:loss}, where $\mathcal{L}_{rcnn}$ can be omitted if there is no RoI head. Refer to \textit{Supp. A.1} for more information.

\begin{equation}
    \label{eq:loss}
    \begin{aligned}
    \mathcal{L}_{det} = \mathcal{L}_{rpn}+\mathcal{L}_{rcnn}+\mathcal{L}_{seg}. \\
    \end{aligned}
\end{equation}

\subsection{Results on WOD}

The results on the \textit{validation} set are displayed in Tab.~\ref{waymoval}, and it can be seen that we achieve new state-of-the-art performance on all the three classes. In particular, for pedestrian, we outperform the baseline model PV-RCNN++ \cite{shi2021pv} by 2.77\%/2.56\% in terms of both L1 and L2 mAP, which indicates the effectiveness of the proposed model in handling hard examples.

\begin{figure}[h!]
    \centering
    \includegraphics[width=0.47\textwidth]{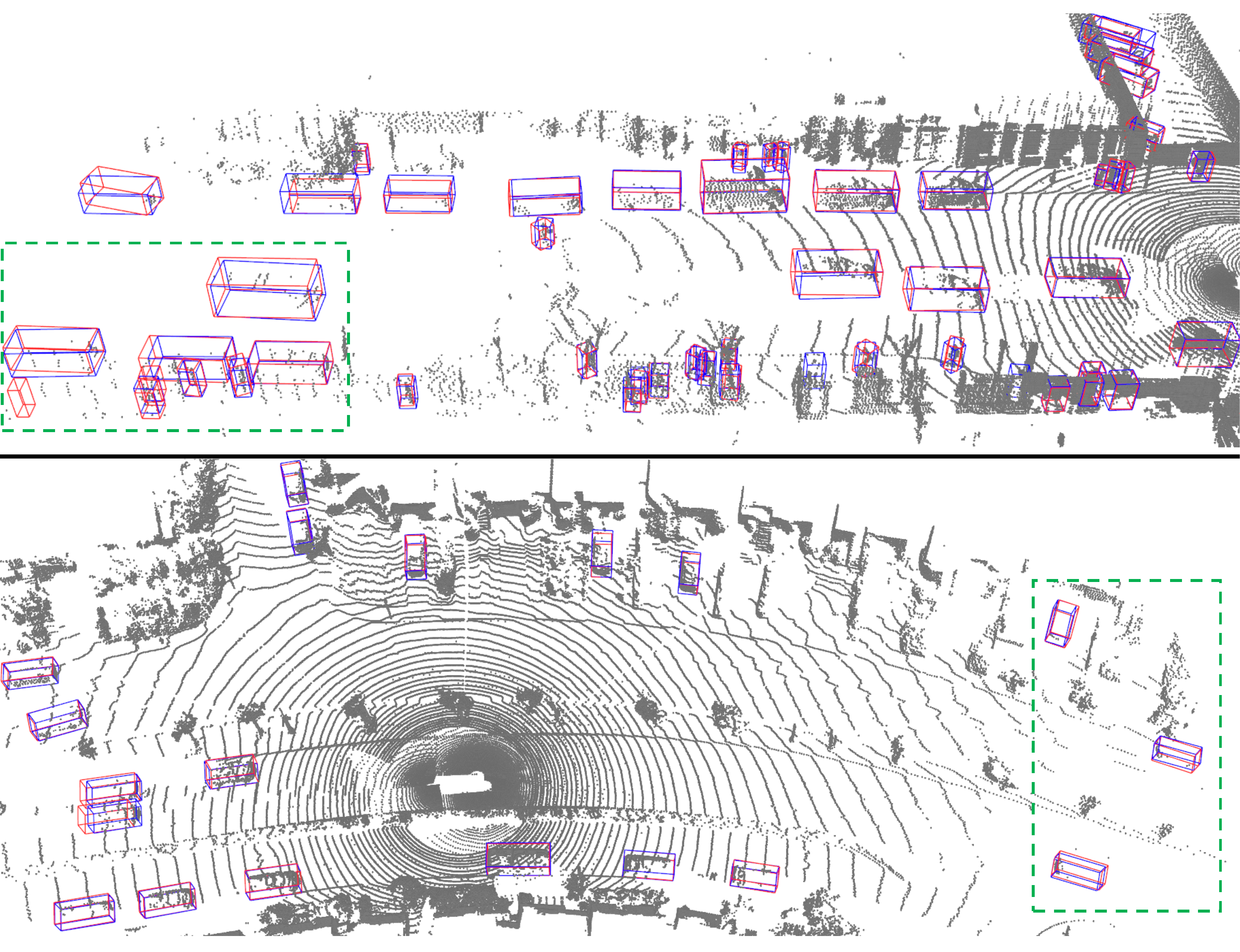}
    \caption{Visualization of results by OcTr on the WOD \textit{validation} split. Blue/red indicates predicted/ground-truth bounding boxes}
    \label{fig_vis}
    \vspace{-0.3cm}
\end{figure}

In comparison with other Transformer-based models, we focus on vehicle since the counterparts only report the performance on it. As Tab.~\ref{waymorange} shows, our OcTr achieves the best mAP among all these convolution- and Transformer-based backbones. It also outperforms the Transformer-based detection head network CT3D \cite{sheng2021improving} by 2.52\% and 1.46\% in L1 and L2 mAP. Regarding the accuracies at different distances, OcTr ranks the first place in the range of 30m-50m and 50m-inf, which surpasses the previous best by 0.45\%, 2.66\% in L1 mAP and 1.18\%, 2.26\% in L2 mAP respectively. It clearly illustrates that OcTr has the advantage in capturing long-range fine-grained context, which facilitates dealing with objects far away. Indeed, far objects generally have much more sparse points than near ones and heavily rely on context for detection, thus benefiting more from OcTr. Refer to Fig.~\ref{fig_vis} for visualization.


\begin{table}[t]
\setlength{\abovecaptionskip}{0.1cm}
    \centering
    \resizebox{0.47\textwidth}{!}{
    \begin{tabular}{c|cc}\toprule[1pt]
    \textbf{Detector} & \textbf{Veh. mAP (L1/L2)} & \textbf{Pedes. mAP (L1/L2)} \\ \midrule
    SECOND\cite{yan2018second} & 70.96/62.58 & 65.23/57.22\\ 
    Ours & \textbf{73.28/65.05} & \textbf{68.08/60.36} \\ \hline
    PV-RCNN\cite{shi2020pv} & 75.41/67.44 & 71.98/63.70 \\
    Ours & \textbf{76.77/68.31} & \textbf{73.22/64.30} \\ \hline
    PV-RCNN++\cite{shi2021pv} & 77.82/69.07 & 77.99/69.92 \\ 
    Ours & \textbf{78.01/69.60} & \textbf{80.75/72.45} \\ 
    \bottomrule
    \end{tabular}
    }
    \caption{Results of extensions to different representative detectors on the WOD \textit{validation} set.}
    \label{ablation_detectors}
    \vspace{-0.5cm}
\end{table}

\subsection{Results on KITTI}

The performance of the single-stage detectors is shown in Tab.~\ref{kittisingle}. We take SECOND\cite{yan2018second}, a commonly used anchor-based model, as the baseline, and compare OcTr with another two advanced Transformer-based variants VoxSet \cite{he2022voxel} and VoTR \cite{mao2021voxel}. We can see that OcTr achieves comparable results with SECOND and VoxSeT in car, while it significantly outperforms all the counterparts and reports the best performance both in pedestrian and cyclist. Benefiting from the large receptive field and fine-grained global context, it exceeds SECOND by 4.19\% and 3.29\% for pedestrians and cyclists respectively, where hard samples often appear.

We summarize the performance of the two-stage models on the KITTI \textit{test} set in Tab.~\ref{kittitwotest}. With the help of the multi-scale backbone features and rich global context, OcTr reaches a leading mAP in car at the moderate level, surpassing the state-of-the-art Focals-Conv \cite{chen2022focal} by 0.36\%. We also evaluate OcTr on the KITTI \textit{val} set, and OcTr again delivers the best performance in the average score, outperforming the second-best by 0.49\%. One can observe that we rank the best or the second best in all the cases.

\subsection{Ablation study}

\begin{table}[t]
    \centering
    \resizebox{0.47\textwidth}{!}{
    \begin{tabular}{c|cc}\toprule[1pt]
    \textbf{Attention} & \textbf{Veh. mAP (L1/L2)} & \textbf{Pedes. mAP (L1/L2)} \\ \midrule
    Ours (\textit{OctAttn}) & \textbf{73.3/65.1} & \textbf{68.1/60.4} \\
    Performer~\cite{choromanski2020rethinking} & 71.4/63.6 & 65.7/57.9 \\ 
    ACT~\cite{zheng2020end} & 71.7/63.5 & 64.3/56.1  \\ 
    VoTr~\cite{mao2021voxel} & 69.4/61.5 & 65.0/57.0 \\ 
    Nearest $K$ & 68.2/59.8 & 64.9/56.7 \\ 
    \bottomrule
    \end{tabular}
    }
    \caption{Ablation on various attention mechanisms and sampling patterns on the WOD \textit{validation} set.}
    \label{ablation_attention}
    \vspace{-0.5cm}
\end{table}

\begin{table}[t]
    \centering
    \resizebox{0.47\textwidth}{!}{
    \begin{tabular}{ccc|cc}\toprule[1pt]
    \textbf{LEPE} &\textbf{SAPE} & \textbf{SAM} & \textbf{Veh. mAP (L1/L2)} & \textbf{Pedes. mAP (L1/L2)} \\ \midrule
    &  &  & 71.35/63.30 & 65.75/57.89\\ 
    \checkmark &  &  & 72.34/64.32 & 66.56/58.62 \\ 
    \checkmark & \checkmark &  & 72.64/64.46 & 66.62/58.83 \\
    \checkmark &  & \checkmark & 72.86/64.40 & 67.79/59.90 \\ 
    \checkmark & \checkmark & \checkmark & \textbf{73.28/65.05} & \textbf{68.08/60.36} \\
    \bottomrule
    \end{tabular}
    }
    \caption{Ablation on semantic positional embedding on the WOD \textit{validation} set.}
    \label{ablation_pe}
    \vspace{-0.1cm}
\end{table}

\begin{table}[t]
    \centering
    \resizebox{0.47\textwidth}{!}{
    \begin{tabular}{c|cc}\toprule[1pt]
    \textbf{top$k$ number} & \textbf{Veh. mAP (L1/L2)} & \textbf{Pedes. mAP (L1/L2)} \\ \midrule
    1 & 70.38/62.20 & 64.19/56.43 \\
    4 & 72.58/64.42 & 66.21/58.42 \\ 
    8 & \textbf{73.28/65.05} & \textbf{68.08/60.36}  \\ 
    16 & 73.25/65.01 & 67.89/60.10 \\ 
    \bottomrule
    \end{tabular}
    }
    \caption{Results of different $k$ values on the WOD \textit{validation} set.}
    \label{ablation_topk}
    \vspace{-0.1cm}
\end{table}

\begin{table}[t]
    \centering
    \resizebox{0.47\textwidth}{!}{
    \begin{tabular}{l|ccc}\toprule
    \textbf{Method} & \textbf{\#Param. (M)} & \textbf{Latency (ms)} & \textbf{Memory (GB)} \\ \midrule
    SECOND\cite{yan2018second} & 5.3  & 48 & \textbf{2.3} \\
    VoTR-SSD\cite{mao2021voxel} & 4.8 & 67 & 3.0 \\ 
    VoxSeT-SSD\cite{he2022voxel} & 3.0 & \textbf{37} &  3.6 \\ 
    OcTr-SSD & \textbf{2.9} & 64 & 2.5 \\ 
    \bottomrule
    \end{tabular}}
    \caption{Resource costs of different backbones with single-stage detectors on the KITTI dataset, test on GTX2080Ti.}
    \label{inference}
    \vspace{-0.1cm}
\end{table}

\noindent\textbf{Scalability on various detectors} As summarized in Tab.~\ref{ablation_detectors}, we conduct experiments on three different and representative detectors, SECOND (single-stage, anchor-based), PV-RCNN (two-stage, anchor-based) and PV-RCNN++ (two-stage, anchor-free). Regardless of the number of stages or region proposal network, we acquire sound improvements compared to the baselines of the sparse convolution backbones, highlighting its scalability.

\noindent\textbf{Ablation on \textit{OctAttn}} We carry out additional experiments to make apple-to-apple comparison with several representative linear Transformer methods, including Performer~\cite{choromanski2020rethinking} (kernel-based linear attention), ACT~\cite{zheng2020end} (cluster-based linear attention), VoTr~\cite{mao2021voxel} (fixed patterns) and the Nearest-$K$ strategy. Tab.~\ref{ablation_attention} lists the results, and our \textit{OctAttn} clearly performs the best, showing its ability.

\noindent\textbf{Ablation on semantic positional embedding} We individually evaluate the contributions of LePE, SAPE and SAM with SECOND as the baseline detector on WOD in Tab.~\ref{ablation_pe}. By incorporating LePE, the L1/L2 performance is boosted by 0.99\%/1.02\% and 0.81\%/0.73\% on vehicle and pedestrian, illustrating its necessity. Furthermore, we separately verify the validity of SAPE and SAM. With semantic clues, we observe that SAPE increases by 0.3\% L1 mAP on vehicle, while SAM provides an L1/L2 mAP improvement of 0.52\%/0.08\% and 1.23\%/1.28\% on vehicle and pedestrian. Finally, we simultaneously apply SAPE and SAM and construct the full model of OcTr, which gains 0.94\%/0.73\% and 1.52\%/1.74\% L1/L2 mAP on vehicle and pedestrian, showing its impact.

\noindent\textbf{Influence by top$k$} Tab.~\ref{ablation_topk} shows the performance of OcTr with various values of $k$. As we can see, the performance improves when $k$ becomes larger but quickly saturates. Due to the redundancy of the scanned scenes, we argue that only a few tokens with high relevance need to be subdivided to embrace fine-grained features.

\noindent\textbf{Analysis on model complexity} We compare OcTr with two recent Transformer-based models in terms of resource cost by keeping the same detection head. Tab.~\ref{inference} shows that with the learnable octree attention mechanism, OcTr consistently maintains less model parameters and less memory occupancies than the counterparts. Regarding the inference speed, VoxSeT runs faster, but it should be noted that VoxSeT inputs with pillars which discard the height dimension, leading to inferior results. As for VoTR, we deliver a mild improvement in efficiency while bringing a large gain in performance.

\section{Conclusion}

This paper proposes a novel voxel-based approach to 3D object detection, namely OcTr. It aims to balance the fine-grained global representation and efficiency with acceptable resource costs. To this end, we propose a learned sparsification attention mechanism, \textit{OctAttn}, which adaptively prunes the octants from the multi-scale feature pyramid in a top-to-bottom manner. Furthermore, we adopt a hybrid semantic-aware positional embedding based on foreground segmentation. Extensive experiments are conducted on WOD and KITTI, and OcTr reaches the state-of-the-art performance, validating its effectiveness.

\section*{Acknowledgment}
This work is partly supported by the National Natural Science Foundation of China (No. 62022011 and No. 62202034), the Research Program of State Key Laboratory of Software Development Environment (SKLSDE-2021ZX-04), and the Fundamental Research Funds for the Central Universities.

{\small
\bibliographystyle{ieee_fullname}
\bibliography{egbib}
}

\clearpage

\setcounter{table}{0}
\setcounter{figure}{0}
\setcounter{section}{0}
\renewcommand\thesection{\Alph{section}}
\renewcommand\thefigure{\Alph{figure}}
\renewcommand\thetable{\Alph{table}}

\noindent \textbf{\Large Supplementary Material} \\

\appendix

This supplementary material provides more implementation details on OcTr in Sec.~\ref{sec:A}, more experiments results in Sec.~\ref{sec:B} and additional visualization in Sec.~\ref{sec:C}.

\section{More Implementation Details}
\label{sec:A}
 
\subsection{Detailed Implementation}

The voxel size in WOD and KITTI is set as [0.1m, 0.1m, 0.1875m] and [0.05m, 0.05m, 0.125m], respectively. The convolution patch embedding module outputs the feature map with a downsampling ratio of 4 and the dimension of the feature map is set as $64 \times 8 \times 376 \times 376$ in WOD and $64 \times 8 \times 400 \times 352$ in KITTI. There are two stacked octree Transformer layers, with two Octree Transformer Blocks (OTBs) in each layer. In the first layer, the pyramid height, the attention dimension, the number of heads, the dimension of heads, and the value of top$k$ are set to 4, 64, 2, 32 and 8, respectively. In the second layer, they are set to 3, 64, 2, 32 and 8. $\tau$ in Eq. \eqref{gumbel} is set to 1 and $\Gamma$ in Eq. \eqref{rpe} is 10000 in practice. During the training procedure, we adopt the Adam optimizer with a batch size of 16, and the cosine annealing learning rate scheduler with an initial value of 0.01 for the two-stage model and 0.003 for the single-stage model. Other hyper-parameters in detection heads, data augmentation and post-processing are set the same as the default values in OpenPCDet \cite{openpcdet2020}.  

Our code is implemented based on OpenPCDet~\cite{openpcdet2020}. All the experiments are conducted on 4 RTX 3090 GPUs except the ones on complexity analysis shown in Tab.~\ref{inference}.

\subsection{Detailed Architecture}

\begin{figure}[t]
    \centering
    \includegraphics[width=0.41\textwidth]{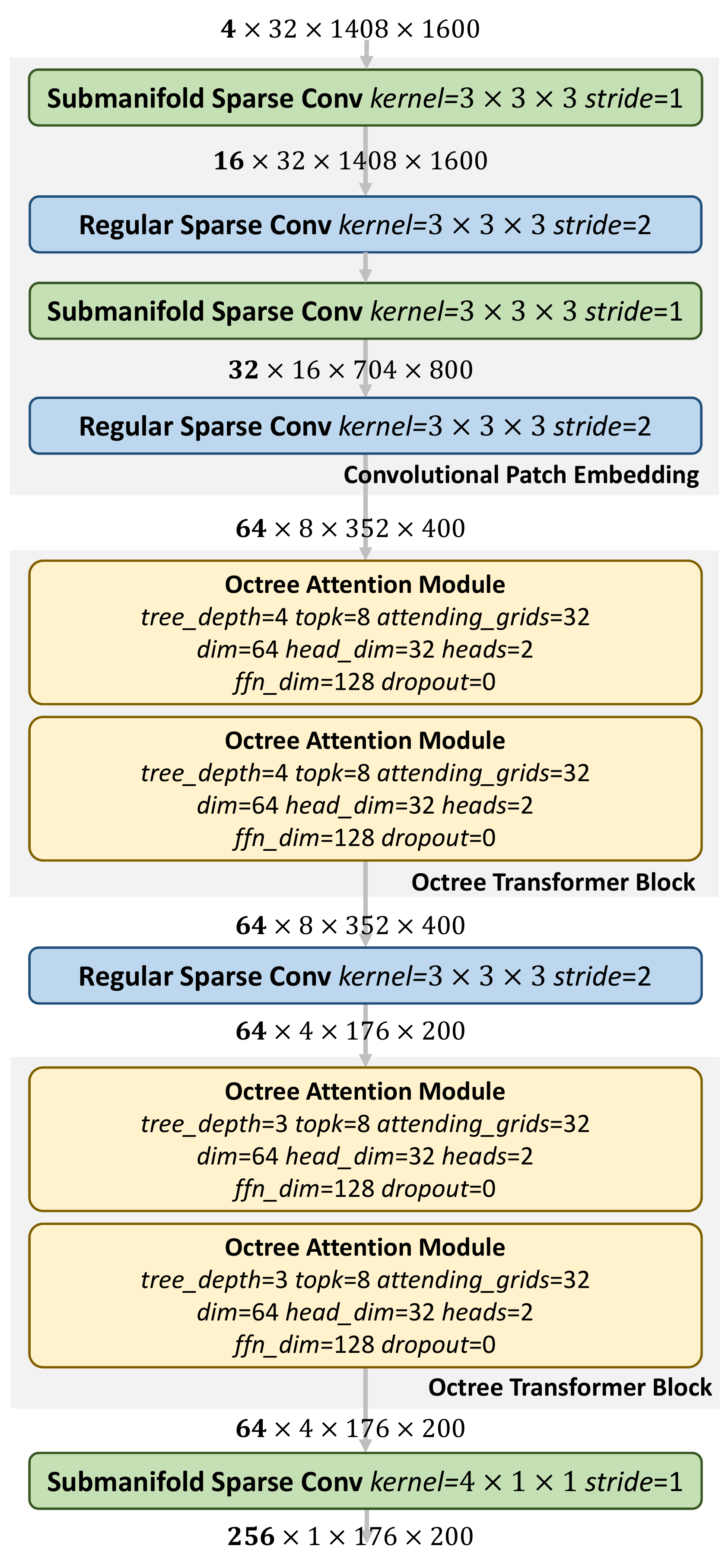}
    \caption{Detailed architecture of the proposed OcTr network.}
    \label{detailed_arch}
    \vspace{-0.5cm}
\end{figure}

The detailed architecture of OcTr is demonstrated in Fig.~\ref{detailed_arch}. The convolutional patch embedding is composed of sparse convolutions with the kernel size of $3\times3\times3$, and 4$\times$ downsampling is conducted on input feature maps. A regular sparse convolution layer is applied for downsampling between OTBs, and the successive height compression operation is replaced with a pixel-wise sub-manifold sparse convolution on BEV features. 

\subsection{Top$k$ Sampling}

\begin{figure}[t]
    \centering
    \includegraphics[width=0.42\textwidth]{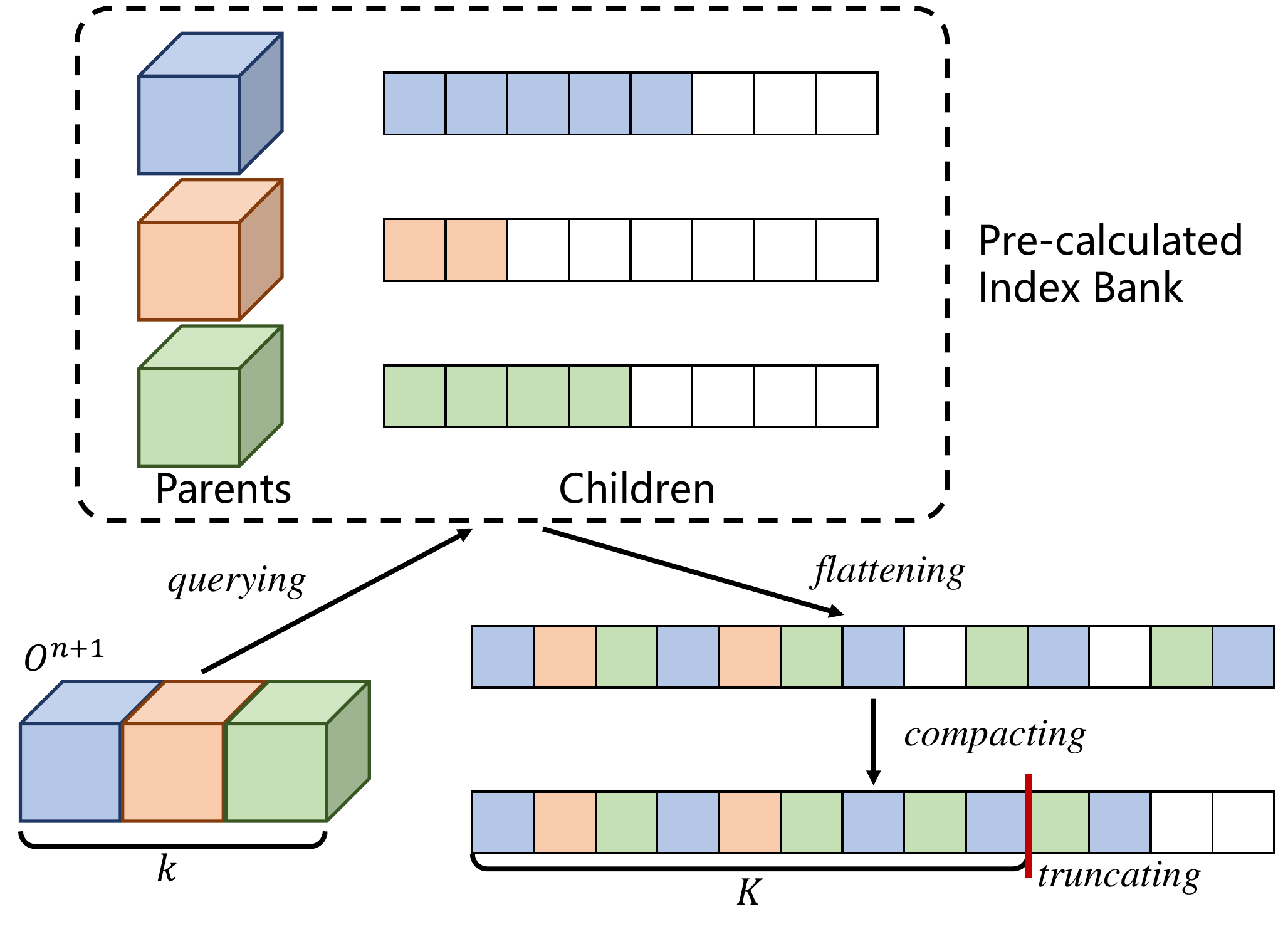}
    \caption{Illustration of top$k$ sampling (the white/colored square denotes empty/non-empty grid, respectively).}
    \label{topk_sampling}
\end{figure}

Top$k$ sampling is an important component in OcTr (in Sec. 3.3 of the main body). To generate the selected sparse octants for subdivision, we first record the indices between the child and parent octants as a pre-calculated index bank. Fig.~\ref{topk_sampling} depicts the entire procedure. In level $n$, by querying about top$k$ parent octants and the pre-calculated index, we densify the sampling outputs $\vec{K} \in \mathbb{R}^{B \times m_{n+1} \times 8\cdot k \times d}$ and flatten the tensors of the key/value in an $8\cdot k \rightarrow [k, 8]$ manner. We then compact the tensor and truncate the top $K$ children, resulting in $\vec{K} \in \mathbb{R}^{B \times m_{n+1} \times K \times d}$. By using the pre-defined index bank, we broadcast the sampled tensors to align the features in layer $n$, generating a tensor with the shape of $\mathbb{R}^{B \times m_{n} \times K \times d}$.

\begin{figure}[t]
    \centering
    \includegraphics[width=0.47\textwidth]{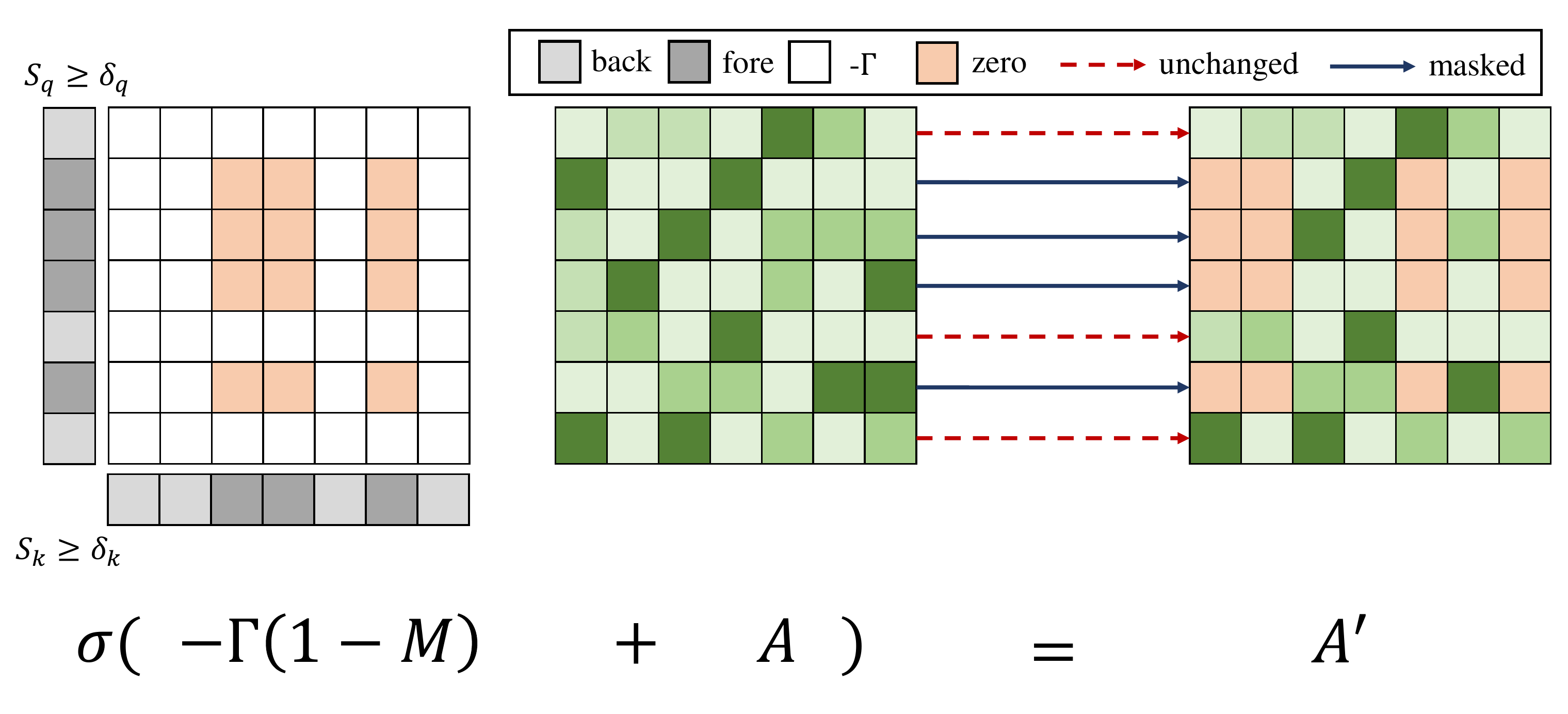}
    \caption{In SAM, the attention scores of background grids maintain unchanged, while those of foreground ones are masked.}
    \label{sam}
\end{figure} 

According to statistics, the downsampling ratio in feature pyramid construction is fixed as 3.2, \emph{i.e.} $m_{n} / m_{n+1} \approx 3.2$. Empirically, to adequately query, we set $K = 4 \times k$ in our implementation.

\subsection{Semantic Attention Mask}

Following Eq. \eqref{rpe} in the main body, we further show the details of SAM in Fig.~\ref{sam}. To obtain a mask for inferior foreground grids, we define a boolean tensor $\mathcal{M}_{q} = \mathbb{I}_{S_{q} \geq \delta_{q}}$, where $S_{q}$ is calculated by the mean scatter function~\footnote{https://pytorch-scatter.readthedocs.io/}. Similarly, we have $\mathcal{M}_{k} = \mathbb{I}_{S_{k} \geq \delta_{k}}$ and obtain the boolean semantic mask on the attention matrices, which is measured by $\mathcal{M} = \mathcal{M}_{q} \cdot \mathcal{M}_{k}$. Though segmentation scores indicate the significance of grids, they suffer from inaccurate predictions. Considering that the mask $\mathcal{M}$ tends to suppress the attention scores of the background grids to 0 and thus deteriorate representations, we simply maintain the attention scores of the background unchanged. The hyper-parameters $\delta_{q}$ and $\delta_{k}$ are set to 0.05 and 0.2, respectively.

\begin{figure}[t]
    \centering
    \includegraphics[width=0.47\textwidth]{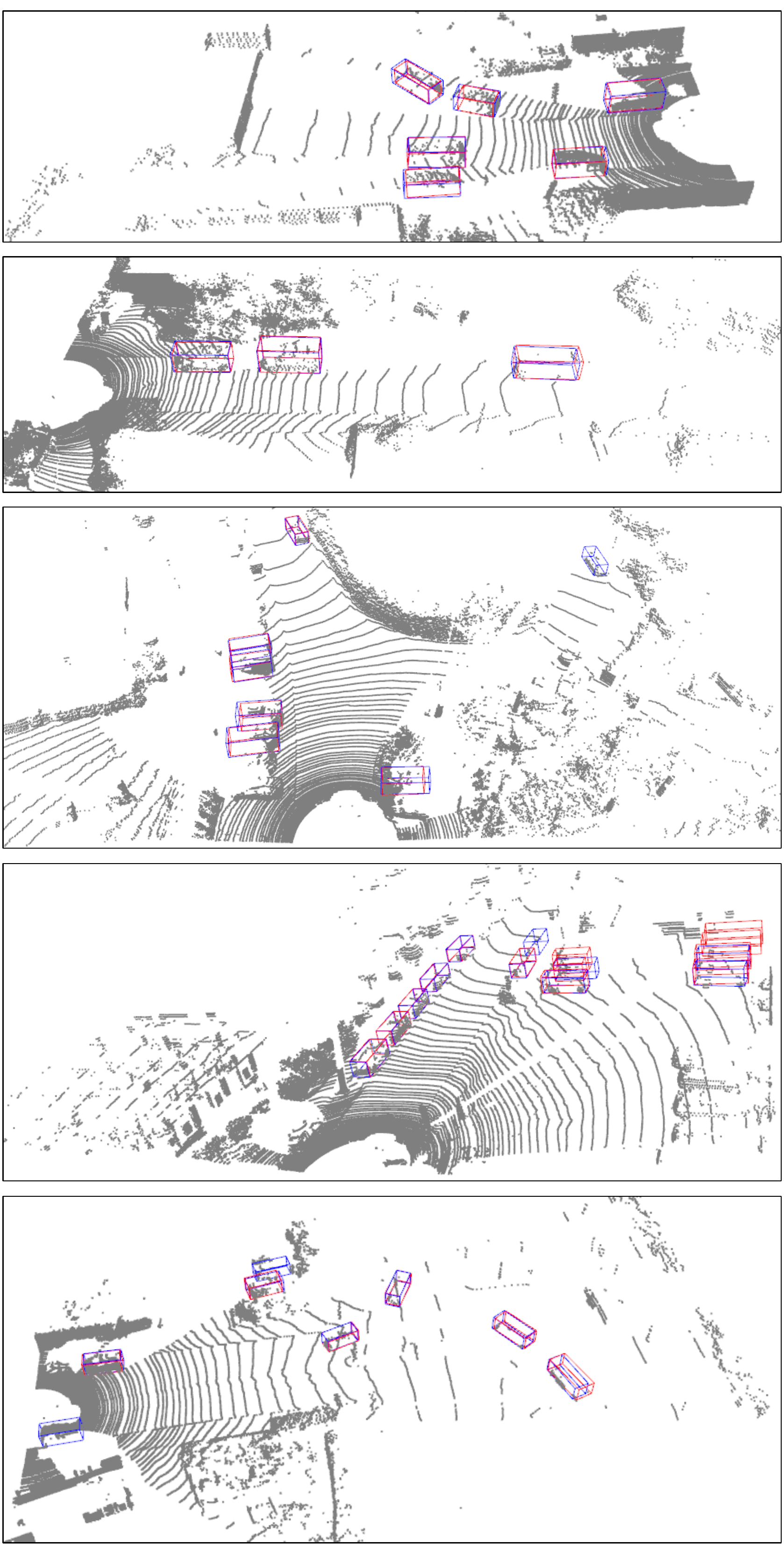}
    \caption{Visualization on the KITTI \textit{val} set. The blue/red bounding boxes indicate the predicted/ground-truth results, respectively.}
    \label{vis_kitti}
\end{figure}

\begin{table*}[t]
    \centering
    \begin{tabular}{l|c|c|c|c|c|c}\toprule[1pt]
    \multicolumn{1}{l|}{\multirow{2}*{\textbf{Model}}} &\multicolumn{1}{c|}{\textbf{Vehicle (L1)}} & \multicolumn{1}{c|}{\textbf{Vehicle (L2)}} &\multicolumn{1}{c|}{\textbf{Pedes. (L1)}} & \multicolumn{1}{c|}{\textbf{Pedes. (L2)}} &\multicolumn{1}{c|}{\textbf{Cyclist (L1)}} & \multicolumn{1}{c}{\textbf{Cyclist (L2)}} \\ 
    ~ & \textbf{mAP/mAPH} & \textbf{mAP/mAPH} & \textbf{mAP/mAPH} & \textbf{mAP/mAPH} & \textbf{mAP/mAPH}  & \textbf{mAP/mAPH} \\ \hline
    CF (1 frame)~\cite{zhou2022centerformer} & 75.2/74.7 & 70.2/69.7 & 78.6/73.0 & 73.6/68.3 & 72.3/71.3 & 69.8/68.8 \\
    CF (8 frames)~\cite{zhou2022centerformer} & 78.8/78.3 & \textbf{74.3/73.8} & 82.1/\textbf{79.3} & \textbf{77.8/75.0} & 75.2/74.4 & 73.2/72.3 \\
    FSD~\cite{fan2022fsd} & 79.2/78.8 & 70.5/70.1 & \textbf{82.6}/77.3 & 73.9/69.1 & \textbf{77.1/76.0} & \textbf{74.4/73.3} \\
    Graph-RCNN~\cite{yang2022graph} & \textbf{80.8/80.3} & 72.6/72.1 & 82.4/76.6 & 74.4/69.0 & 75.3/74.2 & 72.5/71.5 \\
    \hline
    OcTr & 79.2/78.7 & 70.8/70.4 & 82.2/76.3 & 74.0/68.5 & 73.9/72.8 & 71.1/69.2 \\
    \bottomrule
    \end{tabular} 
    \caption{Performance on WOD \textit{validation} with 100\% training data.}
    \label{waymo_full}
    \vspace{-0.3cm}
\end{table*}

\begin{table*}[t]
    \centering
    \begin{tabular}{l|c|c|c|c|c|c}\toprule[1pt]
    \multicolumn{1}{l|}{\multirow{2}*{\textbf{Model}}} &\multicolumn{1}{c|}{\textbf{Vehicle (L1)}} & \multicolumn{1}{c|}{\textbf{Vehicle (L2)}} &\multicolumn{1}{c|}{\textbf{Pedes. (L1)}} & \multicolumn{1}{c|}{\textbf{Pedes. (L2)}} &\multicolumn{1}{c|}{\textbf{Cyclist (L1)}} & \multicolumn{1}{c}{\textbf{Cyclist (L2)}} \\ 
    ~ & \textbf{mAP/mAPH} & \textbf{mAP/mAPH} & \textbf{mAP/mAPH} & \textbf{mAP/mAPH} & \textbf{mAP/mAPH}  & \textbf{mAP/mAPH} \\ \hline
    SECOND\cite{yan2018second} & 76.2/75.7 & 68.3/67.8 & 68.6/55.3 & 62.8/50.5 & 62.4/56.6 & 60.1/54.6 \\
    Ours (SECOND) & \textbf{77.9/77.4} & \textbf{70.2/69.7} & \textbf{71.5/61.1} & \textbf{65.7/56.1} & \textbf{70.7/69.3} & \textbf{68.1/66.8} \\
    \hline
    PV-RCNN++\cite{shi2021pv}
    & 81.6/81.2 & 73.9/73.5 & 80.4/75.0 & 74.1/69.0 & 71.9/70.8 & 69.3/68.2 \\
    Ours (PV-RCNN++) & \textbf{81.7/81.4} & \textbf{74.0/73.6} & \textbf{81.2/75.2} & \textbf{75.0/69.3} & \textbf{73.0/71.8} & \textbf{70.4/69.4} \\
    \bottomrule
    \end{tabular} 
    \caption{Performance on WOD \textit{test} with 100\% training data.}
    \label{waymo_test}
    \vspace{-0.4cm}
\end{table*}

\section{More Experiments Results}
\label{sec:B}

We add experiments with 100\% training data and compare OcTr with Graph-RCNN~\cite{yang2022graph}, FSD~\cite{fan2022fsd} and CenterFormer (CF)~\cite{zhou2022centerformer}. Note that PVRCNN++ in Table 1 is the same as PVRCNN++ (center). As in Table~\ref{waymo_full}, Graph-RCNN and CF (8 frames) achieve higher results, but either with multi-modal or multi-frame data for prediction. When using single frames, OcTr clearly outperforms CF. As for FSD, the performance of OcTr is comparable or even better than that of FSD on vehicle and pedestrian, but is moderately lower on cyclist. However, FSD builds a strong detection head, which tends to be complementary to our OcTr backbone. We believe that OcTr can be further promoted by combining FSD.

We also conduct experiments on Waymo \textit{test} in Table~\ref{waymo_test} using the representative single-stage SECOND and two-stage PVRCNN++, and the results confirm the effectiveness of our method. It should be noted that the common testing tricks, \emph{e.g.} TTA and WBF, are not applied.

\section{More Visualization Results}
\label{sec:C}

\begin{figure*}[t]
    \centering
    \includegraphics[width=0.9\textwidth]{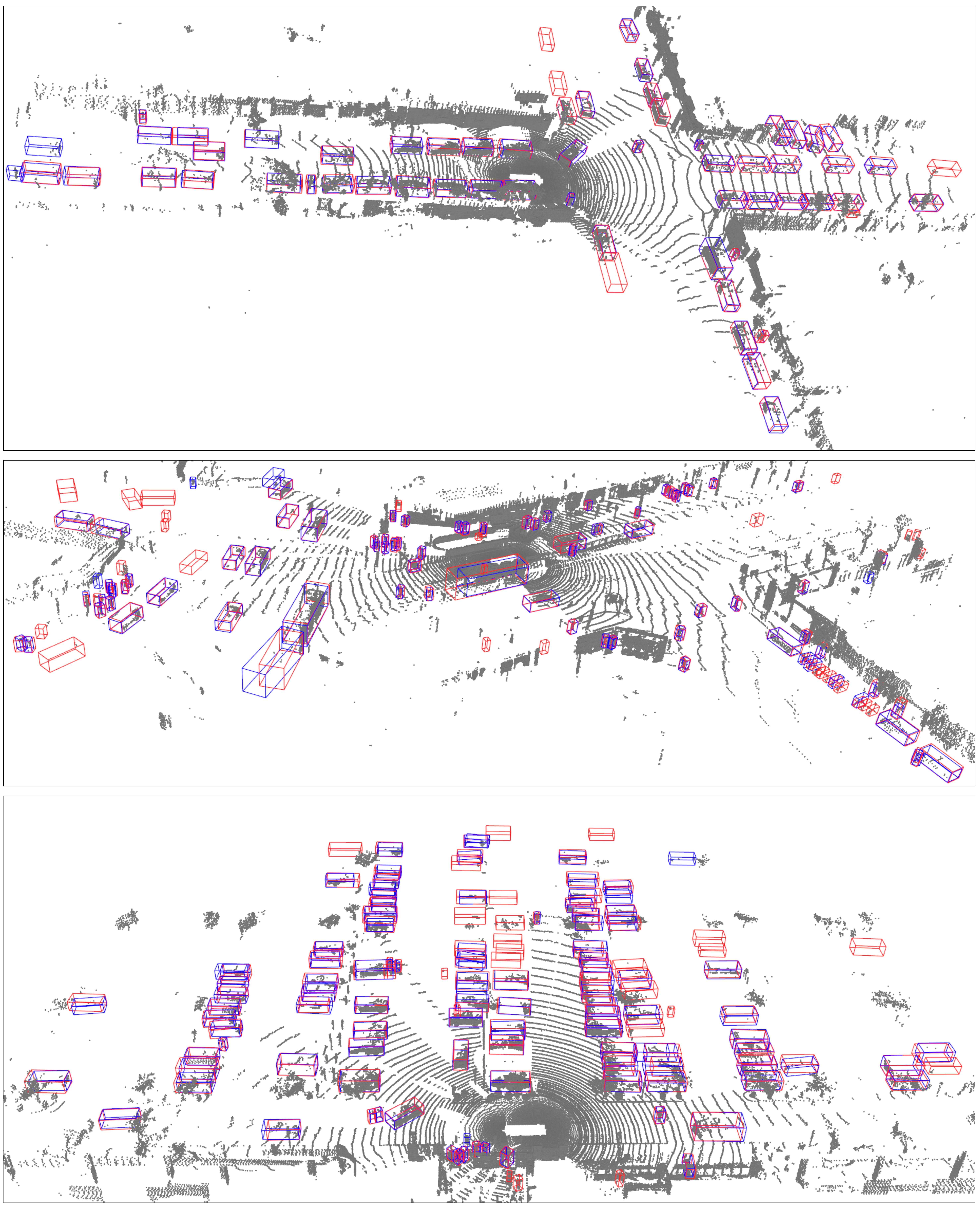}
    \caption{Visualization on the WOD \textit{validation} set in crowded scenes. The blue/red bounding boxes indicate the predicted/ground-truth results, respectively.}
    \label{vis_det_1}
    \vspace{-0.2cm}
\end{figure*}

We additionally visualize some detection results by using the proposed OcTr network on KITTI~\cite{geiger2012we} and WOD~\cite{sun2020scalability} in Fig.~\ref{vis_kitti} and Fig.~\ref{vis_det_1}, respectively. We use the two-stage detector PVRCNN++~\cite{shi2021pv} as our baseline model and only predict cars on KITTI. As displayed, we can observe that OcTr delivers accurate localization and classification for distant and sparse samples, even in crowded scenes.

\end{document}